\documentclass[sigconf]{acmart}
\AtBeginDocument{%
  }

\copyrightyear{2025}
\acmYear{2025}
\setcopyright{acmlicensed}\acmConference[SIGIR '25]{Proceedings of the 48th International ACM SIGIR Conference on Research and Development in Information Retrieval}{July 13--18, 2025}{Padua, Italy}
\acmBooktitle{Proceedings of the 48th International ACM SIGIR Conference on Research and Development in Information Retrieval (SIGIR '25), July 13--18, 2025, Padua, Italy}
\acmDOI{10.1145/3726302.3731944}
\acmISBN{979-8-4007-1592-1/2025/07}

\begin{document}

\title{ADORE: Autonomous Domain-Oriented Relevance Engine for E-commerce}

\author{Zheng Fang}
\email{fangzheng21@jd.com}
\author{Donghao Xie}
\email{xiedonghao1@jd.com}
\affiliation{
  \institution{JD.COM}
  \city{Beijing}
  \country{China}
}

\author{Ming Pang}
\authornote{Authors are corresponding authors.}
\email{pangming8@jd.com}
\author{Chunyuan Yuan}
\authornotemark[1]
\email{chunyuany93@outlook.com}
\author{Xue Jiang}
\email{jiangxue@jd.com}
\affiliation{
  \institution{JD.COM}
  \city{Beijing}
  \country{China}
}

\author{Changping Peng}
\email{pengchangping@jd.com}
\author{Zhangang Lin}
\email{linzhangang@jd.com}
\author{Zheng Luo}
\email{lawching@jd.com}
\affiliation{
  \institution{JD.COM}
  \city{Beijing}
  \country{China}
}

\renewcommand{\shortauthors}{Zheng Fang et al.}

\begin{abstract}
Relevance modeling in e-commerce search remains challenged by semantic gaps in term-matching methods (e.g., BM25) and neural models' reliance on the scarcity of domain-specific hard samples. We propose ADORE, a self-sustaining framework that synergizes three innovations: (1) A Rule-aware Relevance Discrimination module, where a Chain-of-Thought LLM generates intent-aligned training data, refined via Kahneman-Tversky Optimization (KTO) to align with user behavior; (2) An Error-type-aware Data Synthesis module that auto-generates adversarial examples to harden robustness; and (3) A Key-attribute-enhanced Knowledge Distillation module that injects domain-specific attribute hierarchies into a deployable student model. ADORE automates annotation, adversarial generation, and distillation, overcoming data scarcity while enhancing reasoning. Large-scale experiments and online A/B testing verify the effectiveness of ADORE. The framework establishes a new paradigm for resource-efficient, cognitively aligned relevance modeling in industrial applications.
\end{abstract}


\maketitle

\section{Introduction}

E-commerce applications, such as Amazon, Taobao, and JD, have changed people's shopping habits significantly. The relevance module is crucial for the user experience, as displaying irrelevant products can harm trust and revenue for e-commerce systems. Therefore, relevance modeling, which evaluates the match between search queries and products, is a basic part of e-commerce engines.

Traditional relevance models, such as term matching and vector matching methods, have been widely used in online e-commerce systems to ensure fundamental search experience. Term matching based methods apply TF-IDF, BM25~\cite{RobertsonWJHG94} and statistical features to compute the matching degree between query and product, but fail to capture semantic relevance and contextual understanding. Vector matching based approaches~\cite{PangLGXWC16, WanLXGPC16, yuan2024semi, NingPangFang} leverage BERT~\cite{DevlinCLT19} and its variants ~\cite{YaoTCZZY22, YeLZCCWY22, SanthanamKSPZ22, ChenDMJN24, LinTOCYZ24} to understand the semantic relationships between queries and products and achieve remarkable improvement in relevance modeling. 




Despite the progress made, vector matching based methods still face two challenges when tackling domain-specific relevance modeling tasks: (1) The first challenge is the lack of domain-specific hard samples. Most existing methods can achieve satisfactory results by leveraging a moderate amount of manually annotated data. However, further performance improvements often require a substantial number of hard samples, which are not only difficult to obtain but also challenging to annotate accurately. (2) The second challenge is the models' insufficient reasoning and representation capabilities. Constrained by their parameter size, online lightweight models often struggle with complex reasoning and relevance discrimination.  To enhance reasoning and comprehension capabilities, large language models (LLMs) have been increasingly applied to relevance modeling tasks ~\cite{ChenCXLZ23, LiuGJWXG24, 0001SC024, abs-2410-17152, abs-2412-01269, tang2025lref, pang2025generative}. However, most prior LLM approaches primarily utilize the representation of the last token for relevance classification and employ soft labels from the classification layer to distill knowledge into smaller online models. This discriminative paradigm is unable to intuitively analyze which features impact relevance decisions, nor can it enhance the reasoning capabilities of the online student models.

To address the above issues, we propose the Autonomous Domain-Oriented Relevance Engine (ADORE), which consists of three parts: a domain rule-aware relevance discrimination (RD) module, an error-type-aware data synthesis (DS) module, and a key-attribute-enhanced knowledge distillation (KD) module. To automate the construction of domain-specific hard samples, we propose a Chain-of-Thought (CoT) ~\cite{Wei0SBIXCLZ22} based reasoning LLM model within the RD module. To ensure that LLM predictions align with domain rules and user behavior, we use Kahneman-Tversky Optimization (KTO) ~\cite{EthayarajhXMJK24} to align LLM. Furthermore, to endow ADORE with autonomous error correction capabilities, we design an error-type-aware generation model in the DS component. Finally, to effectively transfer the reasoning capabilities of the LLM to the online model, we incorporate the key attributes derived from the CoT analysis into the model input in the KD module, thereby achieving explicit knowledge enhancement. 





In summary, the main contributions of this paper are as follows.
\begin{itemize} 
\item We propose the \textbf{A}utonomous \textbf{D}omain-\textbf{O}riented \textbf{R}elevance \textbf{E}ngine (ADORE), which integrates a domain rule-aware relevance discrimination (RD) module, an error-type-aware data synthesis (DS) module, and a key-attribute-enhanced knowledge distillation (KD) module. 


\item We introduce a CoT-based reasoning LLM in the RD module, which automatically constructs domain-specific hard samples by relevance reasoning on online exposure products. Additionally, the DS module incorporates an error-type-aware generation model to autonomously produce various challenging samples, further enhancing the system's robustness. Key attribute features from CoT analysis are explicitly integrated into the KD module, enabling explicit knowledge enhancement and improving the reasoning and representation capabilities of the online student model.

\item We conduct extensive experiments to evaluate the performance of ADORE, including large-scale offline dataset and online A/B testing. The results demonstrate that ADORE significantly outperforms existing state-of-the-art methods, delivering substantial improvements in both offline and online metrics. This highlights the practical applicability and effectiveness of ADORE in handling complex relevance modeling tasks in e-commerce.
\end{itemize}



\section{METHODOLOGY}

\begin{figure*}[!htbp]
    \centering
    \includegraphics[scale=0.63]{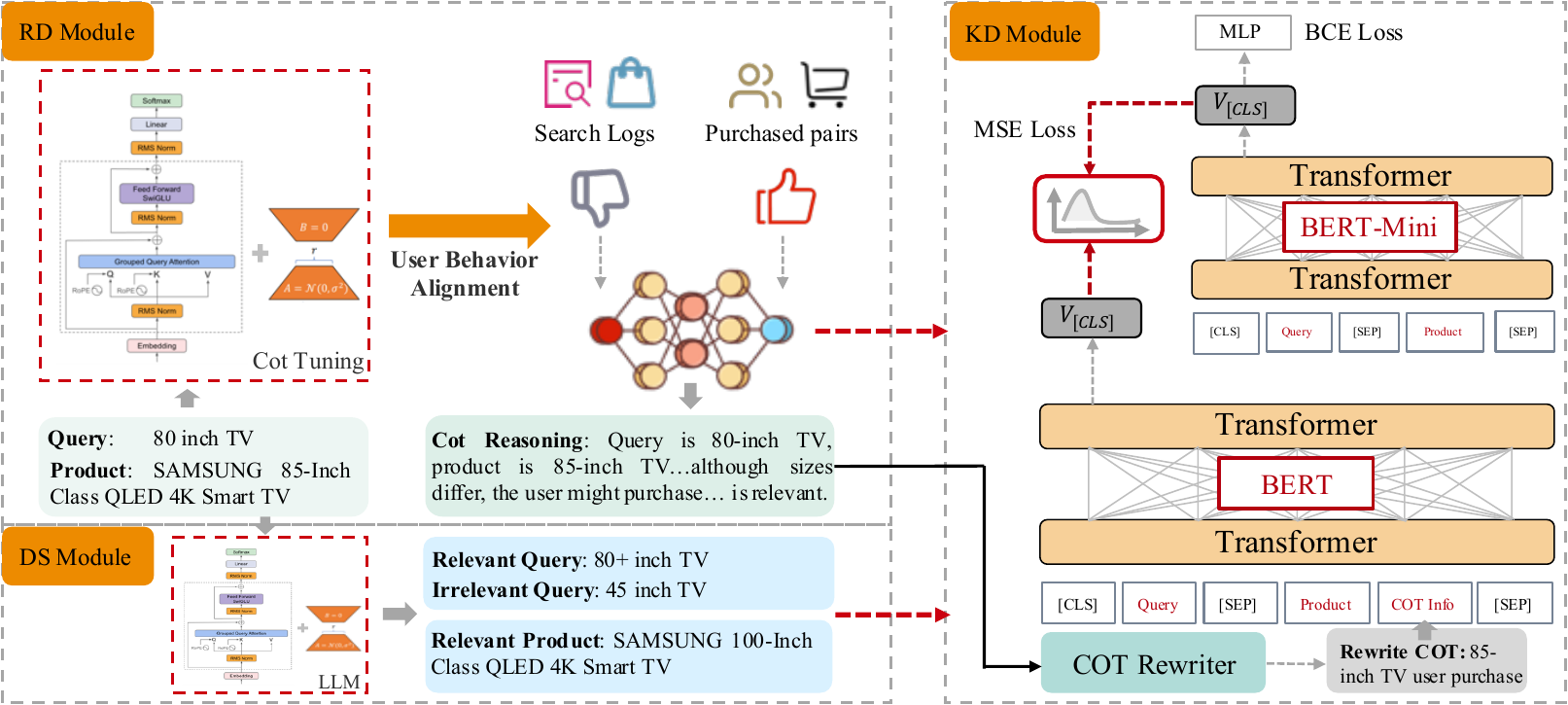}
    \caption{The architecture of the Autonomous Domain-Oriented Relevance Engine.}
    \label{model_structure}
\end{figure*}

In our work, we formulate search relevance prediction as a binary classification problem. Given a query $Q$ and a product $P$, relevance modeling in search engines aims to predict the relevance degree between them. Specifically, the model is designed to classify query-product pairs into one of two distinct relevance categories: Relevant and Irrelevant. Figure~\ref{model_structure} illustrates the components of the Autonomous Domain-Oriented Relevance Engine (ADORE).

\subsection{Rule-aware Relevance Discrimination}
\subsubsection{CoT Tuning}


To identify hard samples that can enhance the performance of our online model, we have developed a relevance discrimination model based on Large Language Models. Previous research ~\cite{Wei0SBIXCLZ22,MadaanHY23} has demonstrated that Chain-of-Thought (CoT) prompting significantly enhances the reasoning accuracy of LLMs. Building on this insight, we fine-tuned our LLM using a CoT-based approach. We initiate the process by utilizing GPT-4o to create CoT labeled data. This is achieved by simulating the human annotation process, where GPT-4o is provided with domain-specific annotation guidelines and a few exemplary CoT instances. As a result, we generate a dataset enriched with detailed reasoning chains, mirroring the depth and nuance of human-like annotation. To further enhance the accuracy of the labeled data, we empirically construct only query-product data that can be validated using human-annotated labels and exclude any CoT data inconsistent with the human labels, forming our final training set. Subsequently, we employ these samples to further fine-tune the LLaMA-3 ~\cite{abs-2407-21783} model through optimization of a standard Language Model loss:
\begin{equation}
\mathcal{L}_{\text{LM}} = -\frac{1}{N} \sum_{i=1}^{N} \log P(w_i | w_{<i})
\end{equation}
The CoT output of the fine-tuned model is illustrated in the figure ~\ref{model_structure}. In this output, the initial step involves extracting key attributes from the query and item descriptions, including product names, brands, model specifications, target audiences, and more. Subsequently, the relevance between the query and item is assessed by evaluating the compatibility of these attributes. Through CoT tuning, the LLM model not only accurately predicts relevance labels but also effectively explains the underlying reasons.



\subsubsection{KTO Alignment}

When analyzing the outputs of the fine-tuned LLM, we observe that it exhibits a bias towards stricter relevance discrimination. Specifically, the model tends to classify products as "Irrelevant" when some attributes mentioned in the query are absent from the product description, even if those attributes are non-essential and do not influence the user's purchasing decision. This tendency towards over-strictness leads to an increase in false negatives, which reduces the model's recall rate and negatively impacts the platform's transaction activity and revenue. To address this issue and ensure that the model's predictions align more closely with real user behavior and feedback, we align the fine-tuned model based on user purchasing behavior. Concretely, we use the fine-tuned model to predict query-purchased item pairs from historical logs and select those that the model classified as "Irrelevant" to form a preference training set. Specifically, we treat these overly strict predictions as loser and use GPT-4o to generate "Relevant" CoT reasoning results for the original query-purchased item pairs as winner. Subsequently, we employ the KTO algorithm to align the LLM with these preferences. In contrast to other alignment algorithms ~\cite{SchulmanWDRK17, RafailovSMMEF23}, the KTO alignment algorithm directly utilizes individual binary labeled data for model alignment, making data construction simpler and more efficient. 



After fine-tuning the LLM and aligning it using the KTO algorithm, we use it as a domain-specific data annotator to label online recalled and exposed items. Hard samples that are inaccurately predicted by the online BERT-based model will be incorporated into the training set,  further improving the online model's performance.



\subsection{Error-type-aware Data Synthesis}


By leveraging the domain rule-aware LLM constructed within the RD module, we can quickly identify hard samples where the online model makes incorrect predictions. However, this piecemeal approach only addresses isolated errors and fails to systematically resolve a particular category of issues. When user search terms or product descriptions undergo subtle changes, the online model may still predict incorrectly. Based on this observation, we have developed an error-type-aware LLM generative model, which automatically synthesizes data that the online model is prone to mispredict, thereby enhancing the model's generalization capability and accuracy. For any given query-product pair, our model can modify the attributes of the original query or product based on instruction prompts, generating new samples that fall into the two categories: Relevant and Irrelevant. Specifically, training samples at this stage are obtained using the GPT-4o model through few-shot prompting. To ensure that the generation model can fully capture the error patterns of the online model,  we iteratively select the generated results where the online model made incorrect predictions as fine-tuning training samples for the LLM. 

After building the error-type-aware LLM-based generative model, we directly sample data from online retrieval and exposed instances, and leverage the model to automatically generate hard samples that are likely to confuse the online model. To mitigate the impact of noisy data on the online model's training, we further utilize the relevance discrimination model from the RD module to filter the generated results.



\subsection{Key-attribute-enhanced Knowledge Distillation}
Given real-time latency and cost considerations, our LLM or deep cross-encoder-based BERT classifier is challenging to scale to the online search system. Therefore, we employ a teacher-student distillation setup to distill a student relevance model, as illustrated in Figure~\ref{model_structure}. Our online model is a lightweight, shallow cross-encoder BERT. To ensure distributional consistency between the teacher model and the student model, we select a deep cross-encoder BERT, which is also encoder-only, as the teacher model, rather than employing an autoregressively trained LLM. Furthermore, to enable the BERT-based model to learn the discriminative reasons for relevance between queries and products, we explicitly incorporate key attribute information mined by the relevance reasoning LLM from the RD module into the input of BERT, aiming for explicit knowledge enhancement. To address the noise and redundancy in lengthy CoT reasoning outputs, we introduce a CoT rewriting model that extracts and refines product attribute descriptions aligned or misaligned with the query. This process filters irrelevant information and reformulates key insights into a concise, structured format, enabling the lightweight BERT-based model to focus on discriminative reasoning. Additionally, since the student model cannot access the CoT results during real-time inference, we only integrate CoT information into the teacher BERT model to ensure consistency between online and offline inputs. Subsequently, the knowledge is transferred to the student model through representation alignment. The loss function is as follows:
\begin{equation}
\mathcal{L}_{\text{kd}}=\alpha \cdot -\sum_{i} y_i \log(\hat{y}_i) + (1 - \alpha) \cdot \frac{1}{H} \sum_{i=1}^{H} (h_i - \hat{h}_i)^2
\end{equation}
where $y_i$ represents the class label, $H$ denotes the dimension size of the $[CLS]$ vector, $h_i$ and $\hat{h}_i$ are the $[CLS]$ representation from the teacher and student model.


\section{Experiment}
\subsection{Datasets and Metrics}

There are no publicly available datasets for the Chinese e-commerce relevance task, we conduct a series of experiments on a large-scale real-world dataset collected from users' search logs in the JD application. This dataset consists of human-annotated query-product pairs, where all annotations are provided by experienced human annotators. The statistics of the datasets are listed in Table~\ref{tab:datset}.

\begin{table}
  \caption{Dataset statistics.}
  \centering
  \label{tab:datset}
  \setlength{\tabcolsep}{1mm}{
      \begin{tabular}{c|ccc}
                \toprule
                \textbf{Statistics} &\textbf{Train} &\textbf{Validation} &\textbf{Test} \\
                \midrule
                \midrule
                \#Query-Product Pairs  & 3,226,439 & 100,000 & 50,961 \\
                \hline 
                \#Uniq. Queries  & 2,862,473 & 87,706  & 45,464\\
                \hline
                \#Uniq. Products  & 3,019,999 & 98,530 & 50,090\\
                \bottomrule
        \end{tabular}
    }
\end{table}

We evaluate ADORE by both offline and online metrics. In offline evaluation, since the human annotations are binary, the task is treated as a classification task, and we evaluate the model by precision, recall, and F1 score. For online evaluation, we conduct experiments in the search advertising and adopt the following business metrics: Ad. CTR (click-through rate), Ad. revenue (total ad revenue), and the Good rate provided by human experts.

\subsection{Baselines}
We compare ADORE with several strong baselines, including widely used BERT, LLAMA-3-8B and their respective variants. For BERT-based models, we compare the two-tower-based BERT ~\cite{HuangHGDAH13} and the cross-encoder-based BERT. For the cross-encoder-based BERT, we further compare its mini version (which shares the same architecture and parameter size as our online model), the standard version, and the version enhanced with our proposed CoT information. As for the LLM model, we compare the approach that directly utilizes the representation of the last token for classification. Finally, we present ADORE, which extends BERT-Mini by integrating online data from the RD module, adversarial-style synthetic data from the DS module, and explicit attribute knowledge from the KD module.


\begin{table}
  \caption{
    The experimental results compared to the baselines on the test dataset. 
  }
  \centering
  \label{tab:experiment}
  \setlength{\tabcolsep}{1mm}{
      \begin{tabular}{l|ccc}
                \toprule
                \textbf{Models} &\textbf{Precision} &\textbf{Recall} &\textbf{F1} \\
                \midrule
                \midrule
                \textbf{DSSM (bi-encoder)}  & 72.72\% & 78.54\%  & 75.52\%  \\
                \textbf{BERT (cross-encoder)} & 84.21\% & 83.93\%  & 84.07\%  \\
                \textbf{BERT-CoT (cross-encoder)}  & 85.35\% & 84.41\%  & 84.88\%  \\
                \textbf{LLM-Classify}  & 84.12\% & 83.28\%  & 83.70\%  \\
                \midrule
                \textbf{BERT-Mini (cross-encoder)}  & 81.10\% & 83.52\%  & 82.29\%  \\
                \textbf{BERT-CoT (+RD\&DS)}  & 86.82\% & 85.56\%  & 86.19\%  \\
                \textbf{ADORE(+RD)}  & 82.61\% & 84.56\%  & 83.57\%  \\
                \textbf{ADORE(+RD\&DS)}  & 84.88\% & 83.96\%  & 84.42\%  \\
                \textbf{ADORE(+RD\&DS\&KD)}  & 85.92\% & 84.98\% & 85.45\%  \\
                \bottomrule
        \end{tabular}
    }
\end{table}

\subsection{Offline Experimental Results}
The experimental results are shown in Table~\ref{tab:experiment}. Overall, the experimental results indicate that ADORE significantly outperforms all baselines on a large-scale real-world dataset. Specifically, we have the following observations: (1) The cross-encoder-based BERT model demonstrates superior performance compared to the bi-encoder-based BERT, and scaling up the model parameters further enhances its performance. Notably, the integration of CoT attributes leads to a marked improvement in BERT's performance, demonstrating the effectiveness of explicit knowledge. Directly using LLM for classification yields results that are even weaker than BERT model. (2) Leveraging the domain-rule-aware LLM within the RD module to generate hard samples significantly enhances model performance. The subsequent inclusion of adversarially crafted data from the DS module further elevates the model's F1 score. Additionally, the explicit knowledge augmentation in the KD module enhances the model's reasoning and generalization prowess, further improving its overall performance.


\subsection{Online Experimental Results}

\begin{table}[!tbp]
  \centering
  \caption{Online improvements of the ADORE. Improvements are statistically significant with $p < 0.05$  on paired t-test. All performances of ADORE and its variants are compared with the online model.}
  \label{tab:online_performance}
  \setlength{\tabcolsep}{1mm}{
      \begin{tabular}{l|cccc}
            \toprule
            Models & \textbf{CTR} & \textbf{Ad. revenue.} & \textbf{Rel.}
            \\
            \midrule
            Online          & - & -  & -   \\
            ADORE(+RD-CoT)  & +0.48\%   & +0.23\%  & +0.50\%   \\
            ADORE(+RD-CoT-Kto)   & +0.98\%   & +1.28\%  & +1.17\%  \\
            ADORE(+RD-CoT-Kto\&DS) & +0.97\%   & +1.07\%  & +1.82\%   \\
            ADORE(+RD-CoT-Kto\&DS\&KD) & +1.24\%   & +1.58\%  & +1.75\%  \\
            \bottomrule
        \end{tabular}
    }
\end{table}

To investigate the effectiveness of our ADORE, we conduct online A/B testing within JD.com's advertising search engine. As shown in Table~\ref{tab:online_performance}, ADORE has delivered significant improvements in human-evaluated relevance. Additionally, it has achieved substantial gains in click-through rate (CTR) and advertising revenue. We can observe that aligning with user behavior and incorporating explicit attribute knowledge results in a significant improvement in user click-through rates, thereby driving effective growth in advertising revenue. Furthermore, with the incorporation of automated mining and synthetic data, online relevance has been significantly enhanced.

\section{Conclusion and Future Work}
In this paper, we present the Autonomous Domain-Oriented Relevance Engine (ADORE), a novel framework designed to address the challenges of scarce domain-specific hard samples and insufficient reasoning capabilities in e-commerce relevance modeling. By leveraging CoT-based reasoning, KTO alignment, adversarial data generation and explicit knowledge-enhanced distillation, ADORE significantly enhances the accuracy, robustness, and reasoning capabilities of online relevance models. Extensive experiments on large-scale real-world datasets and online A/B testing demonstrate that ADORE achieves significant improvements in relevance modeling. The framework has been successfully deployed in JD.com's search system, showcasing its practical applicability and effectiveness in handling complex relevance modeling tasks. For future work, we aim to further enhance ADORE by incorporating multi-modal information, such as product images and user behavior data. Additionally, we plan to explore personalized relevance modeling by integrating user preferences and historical interactions into the framework.

\bibliographystyle{ACM-Reference-Format}
\bibliography{sample-base}

\appendix
\section{Presenter Profiles}
\textbf{Zheng Fang} is a researcher in the Department of Advertising at JD.com Beijing. He received his doctoral degree from the Institute of Information Engineering, Chinese Academy of Sciences. His research focuses on information retrieval. \\
\textbf{Ming Pang} is a researcher in the Department of Advertising at JD.com Beijing. He received his doctoral degree from Nanjing University. His research focuses on machine learning. \\
\textbf{Chunyuan Yuan} is a researcher in the Department of Advertising at JD.com Beijing. He received his doctoral degree from the Institute of Information Engineering, Chinese Academy of Sciences. His research focuses on information retrieval.

\end{document}